\documentclass[acmtog]{acmart}
\acmSubmissionID{1634}
\usepackage[ruled]{algorithm2e} %
\usepackage{float}
\usepackage{enumitem}

\SetAlFnt{\small}
\SetAlCapFnt{\small}
\SetAlCapNameFnt{\small}
\SetAlCapHSkip{0pt}

\usepackage{xcolor}   %
\usepackage{hyperref}
\acmJournal{TOG}

\newcommand{\FP}[1]{{\color{red}{\textbf{Fabio:} #1}}}
\renewcommand{\FP}[1]{}

\newcommand{\revision}[1]{#1}
\usepackage{stfloats}
\usepackage{svg}

\begin{document}

\author[O. El Khalifi]{Omar El Khalifi}
\email{omar@kinetix.tech}
\orcid{0009-0000-1116-4148}
\affiliation{%
  \institution{Kinetix}
  \country{France}
}

\author[T. Rossi]{Thomas Rossi}
\email{thomas.r@kinetix.tech}
\orcid{0009-0009-9650-4783}
\affiliation{%
  \institution{Kinetix}
  \country{France}
}

\author[O. Fossey]{Oscar Fossey}
\email{oscar@kinetix.tech}
\orcid{0009-0008-7426-8604}
\affiliation{%
  \institution{Kinetix}
  \country{France}
}

\author[T. Fouque]{Thibault Fouque}
\email{thibault@kinetix.tech}
\orcid{0009-0003-5296-9366}
\affiliation{%
  \institution{Kinetix}
  \country{France}
}

\author[U. Mizrahi]{Ulysse Mizrahi}
\email{ulysse@kinetix.tech}
\orcid{0009-0005-2795-2321}
\affiliation{%
  \institution{Kinetix}
  \country{France}
}

\author[P. Torr]{Philip Torr}
\email{philip.torr@eng.ox.ac.uk}
\orcid{0009-0006-0259-5732}
\affiliation{%
  \institution{University of Oxford}
  \country{United Kingdom}
}

\author[I. Laptev]{Ivan Laptev}
\email{Ivan.Laptev@mbzuai.ac.ae}
\orcid{0000-0001-7072-3325}
\affiliation{%
  \institution{MBZUAI}
  \country{United Arab Emirates}
}

\author[F. Pizzati]{Fabio Pizzati}
\email{Fabio.pizzati@mbzuai.ac.ae}
\orcid{0000-0002-6249-5649}
\affiliation{%
  \institution{MBZUAI}
  \country{United Arab Emirates}
}

\author[B. Bellot-Gurlet]{Baptiste Bellot-Gurlet}
\email{baptiste.b@kinetix.tech}
\orcid{0009-0005-7062-6783}
\affiliation{%
  \institution{Kinetix}
  \country{France}
}

\title{ActCam: Zero-Shot Joint Camera and 3D Motion Control for Video Generation}

\begin{abstract}
For artistic applications, video generation requires fine-grained control over both performance and cinematography—i.e., the actor’s motion and the camera trajectory. We present ActCam, a zero-shot method for video generation that jointly (i) transfers character motion from a driving video into a new scene and (ii) enables per-frame control of intrinsic and extrinsic camera parameters. ActCam builds on \revision{any} pretrained image-to-video diffusion model that accepts conditioning in terms of scene depth and character pose. Given a source video with a moving character and a target camera motion, ActCam generates pose and depth conditions that remain geometrically consistent across frames. We then run a single sampling process with a two-phase conditioning schedule: early denoising steps condition on both pose and sparse depth to enforce scene structure, after which depth is dropped and pose-only guidance refines high-frequency details without over-constraining the generation. We evaluate ActCam on multiple benchmarks spanning diverse character motions and challenging viewpoint changes. We find that, compared to pose-only control and other pose+camera methods, ActCam improves camera adherence and motion fidelity, and is preferred in human evaluations—especially under large viewpoint changes. Our results highlight that careful camera-consistent conditioning and staged guidance can enable strong joint camera and motion control without training. Project page: \href{https://elkhomar.github.io/actcam/}{https://elkhomar.github.io/actcam/}.

\end{abstract}


\begin{teaserfigure}
\includegraphics[width=\linewidth]{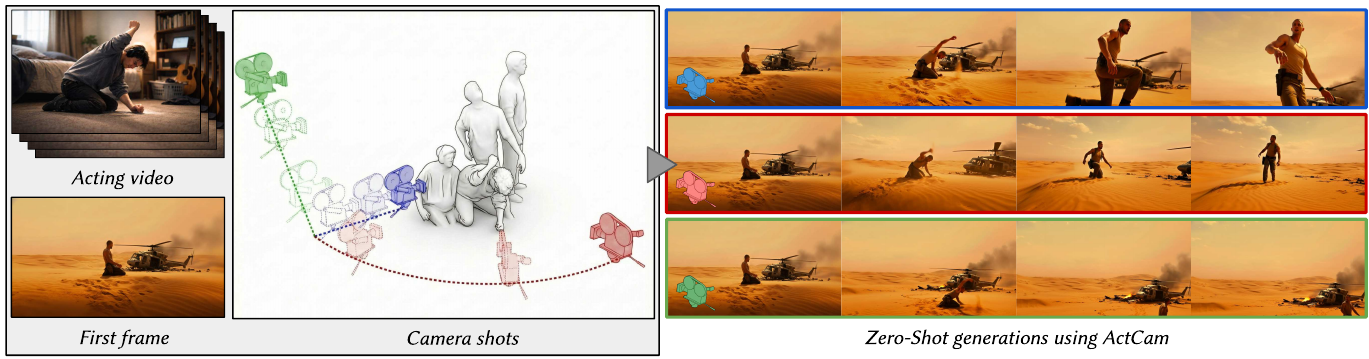}
\caption{\textbf{Overview.} ActCam enables zero-shot joint control of acting motion and camera motion for single-image video generation from a reference image, assuming only widespread conditioning capability of the backbone model on depth and keypoints. Given a reference image, an acting video representing the desired motion, and a target per-frame camera trajectory, ActCam generates a video that preserves identity while following both motion and cinematography.}\label{fig:teaser}

\end{teaserfigure}

\maketitle

\section{Introduction}
\label{sec:intro}

Driven by the intrinsic ambiguities of natural language, video generation has rapidly progressed from text-guided synthesis to strong conditioned generation pipelines~\cite{controlnet,wang2024vace}, capable of accepting additional signals in the generation process to improve fidelity to the user's instructions. This has already opened up possibilities for use in content production. For example, in experimental filmmaking~\cite{elon_musk_reacts_sportskeeda_2025, TangermannFuturismOpenAIHollywood}, visually appealing, consistent shots of synthetic humans can be seamlessly combined to form a short movie. However, results are still far from perfect, mostly due to the limited control over the stylistic characteristics of the output videos. Among other factors, in cinematography, a compelling shot of an actor is defined not only by what the subject does but also by how the camera moves (trajectory, parallax, viewpoint changes). Yet, the capability of controlling both the subject acting performance and the camera movements remains largely unexplored in the current literature.\\ 

To mitigate this problem, a class of motion-control approaches conditions video generation on 2D signals, typically depending on keypoints of human bodies~\cite{wang2025humandreamer,huang2025move}.
While effective under mild viewpoint changes, such controls can become ambiguous under moving cameras: the same 2D signal can correspond to multiple 3D motions, and the control signal may no longer remain consistent once the camera rotates or translates. The closest approach enabling acting videos of humans with a moving camera is Uni3C~\cite{uni3c}, in which 3D representations of humans in motion are combined with camera information thanks to a custom architecture and ad-hoc finetuning. However, we argue that the necessity of using a specific model for generating videos with acting and camera control is a fundamental limitation: First, because the finetuning procedure can be expensive and limitedly transferable to newer models using different architectures. Second, because using a specialized model for this task, potentially combined with other models for shots in which acting is not necessary, may introduce stylistic inconsistencies in the resulting movie, ultimately harming the quality of the end result. \revision{Unlike Uni3C, requiring task-specific training, ActCam operates entirely at inference time and can be applied to any compatible video backbone without modification.}\\

Hence, we introduce \textit{ActCam}, a zero-shot framework exploiting only existing dense 2D conditioning mechanisms, such as depth and human keypoints~\cite{controlnet}, to generate videos of acting humans with 3D camera control. We show some compelling results of ActCam in Figure~\ref{fig:teaser}. Assuming as input an acting video and the first frame of the target scenario, ActCam is able to generate multiple scenarios of the same acting, in the target scene, with arbitrary camera movements, using a pre-trained model with no finetuning. Our main intuition is that the main limiting factor for this task is the current lack of a \emph{camera-aligned} conditioning representation for joint motion and camera control. ActCam addresses this by constructing a multi-modal, camera-aligned conditioning signal, including per-frame pose maps that encode acting motion under the desired camera, and per-frame depth maps that provide a coarse scene geometry proxy under the same camera. For controlling human motion, we benefit from 3D reconstruction of the acting video, avoiding inconsistencies related to the 2D keypoints. For camera control, instead, we benefit from depth reprojection, similarly to related approaches~\cite{uni3c,gen3c}. However, one core characteristic of ActCam is how these conditioning signals interact. To avoid interference between static (depth) and dynamic (human motion) elements in the scene, \textit{we remove the reference character from the scene geometry} and insert the animated character with a novel geometry-aware placement and depth alignment strategy.
Finally, we introduce a \emph{two-phase} conditioning schedule that exploits depth information only in early high-noise steps, while we use only pose information in later steps. This still provides the necessary guidance, removing artifacts due to the over-constraining of the diffusion model with a coarse depth. With both static cameras and dynamic cameras, ActCam outperforms the state-of-the-art under multiple metrics on visual quality and control fidelity. Our contributions are:
\begin{itemize}[topsep=0.5pt]
    \item \textbf{Zero-shot joint control.} We introduce ActCam, a training-free method for joint acting-motion and camera-trajectory control in image-conditioned video generation.
    \item \textbf{Condition construction for joint control.} We design a novel geometry-grounded conditioning pipeline that aligns motion (pose) and scene geometry (depth) to the \emph{target camera} while preventing static/dynamic interference via reference-character removal, geometry-aware placement, and depth alignment.
    \item \textbf{Two-phase conditioning pipeline.} We propose a two-phase inference pipeline that adapts the conditioning information depending on the denoising step, leading to a flexible conditioning that preserves dynamic aspects of the scene.
\end{itemize}

\section{Related Work}
\label{sec:related}

\paragraph{Control signals for video generation.}
A growing body of work improves controllability of diffusion-based image and video generation by injecting external conditions through adapters or control branches, and by conditioning on dense spatial signals such as edges, depth, or semantic layouts \cite{controlnet,animatediff,guo2024i2v,lin2024ctrl}.
These approaches show that dense conditioning can strongly steer generative models. Some focus on motion control, either with reference videos~\cite{pondaven2025video,ling2024motionclone,xiao2024video,yatim2024space} or with additional control signals such as trajectories~\cite{zhang2025tora,yin2023dragnuwa,wu2024draganything,zhou2025trackgo,mou2024revideo} or sparse optical flow~\cite{geng2025motion}. Some impose motion control in a zero-shot manner, although they do not support precise camera control~\cite{burgert2025go}. \citet{matchdiffusion} manipulate noise at inference time to generate videos with similar motion. However, none of those methods solve the joint requirement of controlling an articulated performance while enforcing a non-trivial camera trajectory.

\paragraph{Human-oriented video generation}
Recent reference-based animation methods condition video diffusion on human-centric signals (e.g., keypoints, pose, dense pose) to reproduce an intended performance while preserving identity \cite{moore,humanvid,mimicmotion,animatex,hypermotion,unianimate,wang2024vace,wananimate,steadydancer}. \citet{wang2025humandreamer} uses keypoints as intermediate condition to render more realistic videos starting from text. \citet{xu2024magicanimate} exploits dense human part masks for guiding generation. All these approaches may suffer from occlusions and perspective-induced errors. Recent efforts~\cite{gan2025humandit} have also focused on increasing the length of synthesized human videos. Differently, some use reconstructed 3D humans to condition video generation~\cite{zhu2024champ,liang2025realismotion}, although they offer no camera control. \revision{Similarly, \citet{kulal2023affordance} insert humans into scenes by inferring affordance-aware poses, but without motion or camera control.} Our proposal, instead, offers unified camera control and human motion conditioning by exploiting 3D humans.

\paragraph{Camera control and joint camera--motion control.}
There is an interest in controlling camera trajectories in generated videos. A first line of work exploits Plucker embeddings and additional trained branches to enforce camera control~\cite{kuang2024collaborative,he2024cameractrl,xu2024camco,bahmani2024vd3d,bahmani2025ac3d}. Alternatively, camera trajectories can be enforced by geometry-aware conditions that convey viewpoint changes more directly \cite{gen3c,hou2024training,mengnvs}. These systems do not allow for joint human and camera conditioning simultaneously.
\revision{Concurrently, Pulp Motion~\cite{pulpmotion} generates camera trajectories and human motions, but focuses on trajectory generation rather than video synthesis.}
The closest work to ours is Uni3C~\cite{uni3c}, that allows joint camera and human motion control. However, it relies on additional training/finetuning to unify camera and motion control. In contrast, ActCam outperforms Uni3C exploiting joint control by constructing camera-aligned pose and depth conditions from a reference image, an acting video, and a camera preset.

\section{Method}
\label{sec:method}

\begin{figure*}[t]
  \centering
  \includegraphics[width=\textwidth]{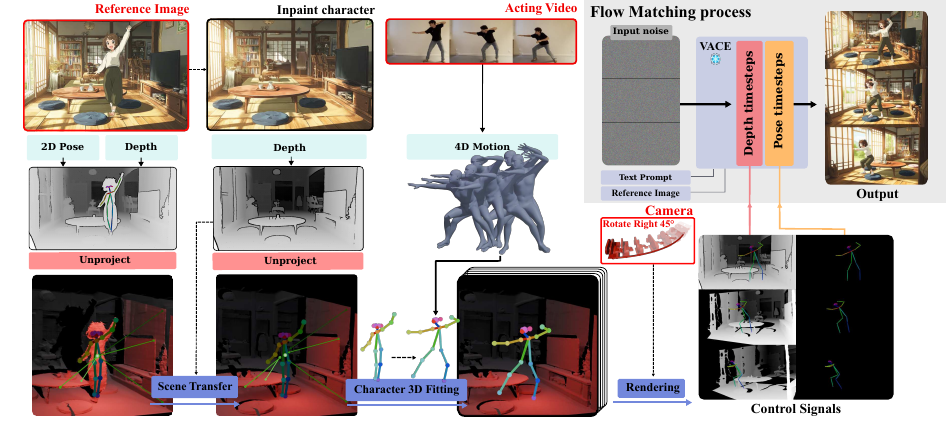}
  \caption{\textbf{ActCam pipeline.} Given a reference image, an acting video, and a target camera trajectory, we (1) estimate background depth from an inpainted reference, (2) recover motion and align it to the background scene via fitting, and (3) rasterize pose and depth+pose control signals under the target viewpoint. A two-phase denoising schedule conditions early steps on
  depth+pose for stronger camera control, then refines with pose-only to encourage motion adherence.}
  \label{fig:pipeline}
\end{figure*}

\subsection{Problem setup}
\label{sec:method:setup}

\paragraph{Inputs and goals.}
We study image-conditioned video generation with joint control over the character's motion and the camera trajectory, using a pretrained conditioned video diffusion backbone. We assume a sequence length $T$ and three inputs: a reference image $I_{\mathrm{ref}} \in \mathbb{R}^{H\times W\times 3}$ defining the target character identity and the environment appearance,
 an acting video $V_{\mathrm{act}}=\{I^{\mathrm{act}}_\tau\}_{\tau=1}^{T}$ providing the target performance, and
 a target per-frame camera trajectory $\mathcal{C}=\{(K_\tau,R_\tau,\mathbf{t}_\tau)\}_{\tau=1}^{T}$, with intrinsics $K_\tau\in\mathbb{R}^{3\times 3}$ and extrinsics $(R_\tau,\mathbf{t}_\tau)\in SO(3)\times\mathbb{R}^3$. Let $V=\{I_\tau\}_{\tau=1}^{T}$ be the generated video.
Our goal is to generate $V$ such that the identity and appearance match $I_{\mathrm{ref}}$, the motion in the output video matches the performance in $V_{\mathrm{act}}$, and finally, such that each frame's viewpoint is in accordance with the desired camera parameters $\mathcal{C}$.

\paragraph{Formalisation of the setup.}
We formalize the pretrained VACE~\cite{wang2024vace} model as a mapping $f: \mathfrak{C} \rightarrow \mathcal{V}$ that transforms a set of dense video-aligned signals $C$ into a video sequence $V$.
VACE is trained on a video dataset combining various video-aligned conditions such as: depth, optical flow, character poses, and activation masks. We restrict the conditioning input to the tuple $C = (I_{\mathrm{ref}}, c_{\mathrm{pose+depth}}, c_{\mathrm{pose}})$. Our objective is to synthesize novel, geometrically consistent dense conditions $c_{\mathrm{pose}}$ and $c_{\mathrm{depth}}$ derived jointly from the acting video $V_{\mathrm{act}}$ and the target camera $\mathcal{C}$, such that the generated video $V = f(C)$ satisfies the desired motion and trajectory constraints. ActCam is a pure inference-time method: we keep the backbone fixed and design conditioning signals that jointly encode motion and camera.

\paragraph{Continuous flow formulation.}
Let $z_t$ denote the latent video state at continuous time $t \in [0, 1]$ and let $C$ denote the conditioning input. We model the generative dynamics as a flow governed by an ordinary differential equation (ODE). The backbone network $v_\theta$ approximates the instantaneous velocity field, defining the temporal evolution of the latent state as
\begin{equation}
    \frac{\mathrm{d}z_t}{\mathrm{d}t} = v_\theta(z_t, t, C).
    \label{eq:flow_ode}
\end{equation}
The target latent $z_1$ is obtained by integrating this flow over time from $ z_0 \sim \mathcal{N}(0, 1)$ and is subsequently decoded into the video $V$. 

\paragraph{Conditioning Requirements for Camera-Aligned Motion Control.}

To jointly ensure camera control and pose reproduction, the conditioning must provide \emph{spatially grounded, per-frame cues in the target camera view}: it must both enforce the intended motion and stabilize the scene layout under viewpoint changes. To enforce camera control, we preferred a dense per frame aligned depth signal making sure the model understands the 3D geometry and camera motion rather than numerical camera control using Plücker rays since dense superiority has been proven in \cite{gen3c}. 
In practice, simply combining off-the-shelf controls is brittle: view-locked motion cues (e.g., 2D keypoints) become ambiguous under camera motion, and conditioning on depth estimated from $I_{\mathrm{ref}}$ entangles the \emph{static} reference character with the \emph{dynamic} character we want to animate, which can cause duplicated characters, freezing, or violations of motion/camera constraints.
ActCam addresses this by constructing camera-aligned pose/depth conditions and by using a two-phase conditioning schedule (Sections~\ref{sec:method:conditions}--\ref{sec:method:two_phase}).

\subsection{Camera-aligned condition construction}
\label{sec:method:conditions}

\paragraph{Intuition.} Our goal is to provide the diffusion model with a target-view-aligned condition that jointly encodes character motion and camera motion. To this end, we leverage depth as a geometric prior: once rasterized under the target camera 
trajectory, the depth signal implicitly conveys viewpoint changes as apparent background motion. However, naively using the reference image depth introduces a static character that conflicts with the dynamic pose signal (see Figure \ref{fig:pipeline}). We address this by inpainting   
the character out of the reference image and estimating a background-only depth map $\mathcal{D}_{\text{bg}}$. ActCam then constructs a unified 3D scene from $\mathcal{D}_{\text{bg}}$ and the recovered poses, from which we rasterize two control signals under the target    
camera: a depth+pose condition and a pose-only condition obtained by omitting the depth.

\paragraph{Camera motion and 3D anchor}
In order to provide a 3D anchor to VACE, we construct a 3D background environment which we argue, once rasterized, will provide enough information to the model to effectively encode the camera motion as an equivalent background motion, and disentangle the character motion from camera motion, reducing conflicts between the two signals. We estimate a depth map $D_{\mathrm{ref}}$ from $I_{\mathrm{ref}}$ using a monocular depth estimator (MoGe \cite{moge}). This leads to the creation of a 3D mesh that can be rendered from different viewpoints. Mesh-based rendering provides stronger geometric consistency than point clouds, which tend to exhibit sparsity and visual artifacts under camera motion.

Nonetheless, solely relying on the reference image depth proves insufficient. Indeed, the presence of a static character in the 3D mesh provides a conflicting signal when rendered with a dynamic pose control signal, as in Figure~\ref{fig:pipeline}. To tackle this, we propose to inpaint the character out of the reference image $I_{\mathrm{ref}}$ to extract a \emph{background-only} depth map $\mathcal{D}_{\text{bg}}$, yielding a \textit{background 3D mesh}, suppressing depth-static character related issues, such as character duplication in the output video.

Also, since VACE relies on dense conditioning, enforcing strict pixel correspondence between the control signals and the generated video is crucial. A key challenge is the alignment of the dynamic actor pose with the background 3D scene geometry. In practice, the reference depth map $\mathcal{D}_{\mathrm{ref}}$ and the background depth map $\mathcal{D}_{\mathrm{bg}}$ are estimated in two independent passes, leading to inconsistencies in 3D space. To resolve this discrepancy, we estimate a geometric alignment allowing the character to be correctly registered within the background 3D mesh. We refer to this stage as the \textit{scene transfer}, described in the next paragraph, and visible in Figure~\ref{fig:pipeline}.

\paragraph{Scene transfer}
Let $\mathcal{M} \subset \llbracket 1,H \rrbracket \times \llbracket 1,W \rrbracket$ denote the binary character segmentation mask in $\mathcal{I}_{\mathrm{ref}}$, obtained using \cite{liu2024grounding}. Our goal is to find the new position of the reference character when \textit{transferred} into the 3D background mesh, while preserving pixel reprojection with $\mathcal{I}_{\mathrm{ref}}$; therefore, only the depth of the character points is adjusted, since its scale and image plane coordinates will be imposed. We exploit the set of non-inpainted pixels $(u,v) \notin \mathcal{M}$, for which reliable correspondences exist between the two depth maps. Let $\mathbf{x}^{\mathrm{ref}}_{u,v} \in \mathbb{R}^3$ and $\mathbf{x}^{\mathrm{bg}}_{u,v} \in \mathbb{R}^3$ denote the 3D points reconstructed from $\mathcal{D}_{\mathrm{ref}}$ and $\mathcal{D}_{\mathrm{bg}}$, respectively. To emphasize constraints near the character boundary, we assign each an importance weight\revision{: not all environment points contribute equally to the scene transfer, as points closer to the character are more informative for rendering correct scene interactions. We assign higher weight to closer points:}
\begin{equation}
w(u,v) = \exp\Big(-\mathrm{dist}(\mathbf{x}^{\mathrm{ref}}_{u,v}, \mathcal{M})\Big),
\end{equation}
where $\mathrm{dist}(\cdot, \mathcal{M})$ denotes the Euclidean distance in the image plane to the closest pixel in the mask $\mathcal{M}$. We then compute the weighted centroids of the character relative to each set of points in $\mathcal{D}_{\mathrm{ref}}$ and $\mathcal{D}_{\mathrm{bg}}$ respectively:
\begin{equation}
\mathbf{p}_{\mathrm{ref}} = \frac{\sum_{(u,v) \notin \mathcal{M}} w(u,v) \, \mathbf{x}^{\mathrm{ref}}_{u,v}}{\sum_{(u,v) \notin \mathcal{M}} w(u,v)}, \qquad
\mathbf{p}_{\mathrm{bg}} = \frac{\sum_{(u,v) \notin \mathcal{M}} w(u,v) \, \mathbf{x}^{\mathrm{bg}}_{u,v}}{\sum_{(u,v) \notin \mathcal{M}} w(u,v)}.
\end{equation}
Assuming that the relative position of the character with respect to these centroids is preserved up to a global depth scaling, we align the character by applying an affine transformation along the depth axis. For any character point with reference depth $z^{\mathrm{char}}_{\mathrm{ref}}$, the aligned depth in the background coordinate system is given by
\begin{equation}
z^{\mathrm{char}}_{\mathrm{bg}} = 
\big(z^{\mathrm{char}}_{\mathrm{ref}} - p^z_{\mathrm{ref}}\big) \frac{p^z_{\mathrm{bg}}}{p^z_{\mathrm{ref}}} + p^z_{\mathrm{bg}},
\end{equation}

where $p^z_{\mathrm{ref}}$ and $p^z_{\mathrm{bg}}$ denote the $z$-coordinates of $\mathbf{p}_{\mathrm{ref}}$ and $\mathbf{p}_{\mathrm{bg}}$, respectively. This transformation jointly accounts for scale and translation mismatches, enabling consistent addition of the character points to the background 3D scene, while effectively taking character-environment proximities (\textit{e.g.} contacts) into account via the importance weighting.

\paragraph{4D motion recovery (acting).}
We recover a motion sequence of 3D humans from $V_{\mathrm{act}}$ using a monocular 3D human motion estimator (GVHMR \cite{gvhmr}).
We denote the recovered articulated state at frame $\tau$ as $\mathcal{S}_\tau$ (e.g., SMPL parameters and root pose).
Unlike 2D keypoints, $\{\mathcal{S}_\tau\}$ reduces depth ambiguity and provides a stable motion signal under viewpoint changes.

\paragraph{Character 3D fitting}
As in \cite{uni3c}, we align the dynamic poses $\mathcal{S}$ with the actual replaced character in the assembled background 3D scene (see Figure \ref{fig:pipeline}). To that extent, we adopt the same approach as in \cite{uni3c} using a least squares estimation based on rigid transformation at time $\tau=0$ between the $\mathcal{S}_0$ and the extracted keypoints from $I_{\mathrm{ref}}$. This method \cite{88573} provides a rotation matrix $R \in SO(3)$, a translation $\hat{t}\in \mathbb{R}^{3}$ and a scale $s$ that we will apply to the pose sequence as follows: 
$$\hat{\mathcal{S}}_{\tau} = s.R\mathcal{S}_{\tau}  + \hat{t}$$
This new sequence $\hat{\mathcal{S}}_{\tau}$ is the one being rendered.

\paragraph{Rendering control signals} 
VACE control signal is a \emph{dense, image-like} representation that can be directly processed by standard video encoders. We thus directly rasterize both the \emph{pose} and the \emph{depth+pose} control signals as videos. We denote $\mathcal{R}^{\mathcal{C}}$ the rasterization operator under the camera $\mathcal{C}$. For the \emph{pose}, we rasterize the animated skeleton $\hat{\mathcal{S}}_{\tau}$ following the standard openpose representation in other works (see \cite{wang2024vace,steadydancer}) encoding 2D joint locations and limb connectivity. This yields a pose control video $c_{\mathrm{pose}} = \mathcal{R}^{\mathcal{C}}(\hat{\mathcal{S}}_{\tau}) \in \mathbb{R}^{T\times H \times W \times 3}$, rendered on a black background, following the target camera viewpoint (see Figure \ref{fig:pipeline}). For the \emph{depth+pose} control signal, we simply render the \emph{background} 3D mesh built from $\mathcal{D}_{\mathrm{bg}}$ using min-max depth normalized grayscale color values, following the format that VACE uses at train time. To obtain actual \emph{depth+pose} joint signal, we finally superimpose the previously rendered pose signal with the \emph{background} depth rasterization to obtain $c_{\mathrm{pose+depth}} = \mathcal{R}^{\mathcal{C}}(\hat{\mathcal{S}}_{\tau}, \mathcal{D}_{\mathrm{bg}}) \in \mathbb{R}^{T\times H \times W \times 3}$.

\subsection{Two-phase conditioning}
\label{sec:method:two_phase}

Monocular depth estimates are often coarse and locally inaccurate. Conditioning solely on actor poses fails to capture camera motion, as the model interprets such motion as part of the body rather than the camera. On the other hand, conditioning on both depth and poses throughout all denoising steps can over-constrain the generation, leading to static backgrounds and the propagation of depth artifacts into high-frequency details (see Ablation, \ref{fig:ablation_depth_only}). To address this, we adopt a two-phase conditioning schedule that gradually relaxes depth conditioning.

Let $t$ denote the diffusion timestep. We define the conditioning signal $c(t)_\tau$ using a cutoff threshold $t_{\text{stop}}$:

\begin{equation}
c(t)_\tau =
\begin{cases}
\mathcal{R}^{\mathcal{C}_{\tau}}(\hat{\mathcal{S}}_{\tau}, \mathcal{D}_{\mathrm{bg}}), & \text{if } t \le t_{\text{stop}}, \\
\mathcal{R}^{\mathcal{C}_{\tau}}(\hat{\mathcal{S}}_{\tau}), & \text{if } t > t_{\text{stop}}.
\end{cases}
\label{eq:two_phase_t}
\end{equation}
The pose signal is derived from 3D motion recovery and target-camera alignment, and provides a stable motion constraint.
We therefore keep pose conditioning throughout the full denoising process.

\section{Experiments}
\label{sec:experiments}

\subsection{Setup}
\label{sec:exp_setup}

We evaluate ActCam having an image $I_{\mathrm{ref}}$, an acting video $V_{\mathrm{act}}$, and optionally a target per-frame camera trajectory as inputs. We adopt VACE~\cite{wang2024vace} as backbone. We set the same resolution ($H{\times}W$)\FP{Which one}, number of steps ($N$)\FP{Which one}, and scheduler\FP{Which one} across methods.%

\paragraph{Datasets.}
For moving camera, we design an evaluation inspired by Uni3C~\cite{uni3c}. We select 4 camera presets with common cinematic motions and evaluate on 100 reference clips from RealisDance-Val~\cite{realisdance} per preset (total 4$\times$100 tests). For each clip, we sample a reference $I_{\mathrm{ref}}$, extract an acting-motion signal from the original clip, and generate a new video under one of the 4 camera presets. For baselines, we input the same setup, meaning same $I_{\mathrm{ref}}$, extracted motion, camera preset. For static cameras, we use RealisDance-Val~\cite{realisdance} under fixed viewpoint.

\paragraph{Baselines.}
For moving camera, we compare with Uni3C~\cite{uni3c}, the most similar method in literature. \revision{Both ActCam and Uni3C are based on the same Wan 2.1 14B backbone, ensuring a fair comparison.} We also test ActCam in a static camera setup by comparing against strong motion control methods without explicit camera control: Moore-AnimateAnyone \cite{moore}, HumanVid~\cite{humanvid}, MimicMotion~\cite{mimicmotion}, Animate-X~\cite{animatex}, Hyper-Motion~\cite{hypermotion}, UniAnimate-DiT~\cite{unianimate}, VACE~\cite{wang2024vace}, Wan-Animate~\cite{wananimate}, and SteadyDancer~\cite{steadydancer}.  We intentionally do \emph{not} compare to camera-control-only methods, as our goal is joint control with motion.
\paragraph{Metrics.}
We follow Uni3C~\cite{uni3c} for the evaluation protocol. First, we use VBench~\cite{vbench} for evaluating the visual quality of the generated videos. We evaluate Subject Consistency (SC), Background Consistency (BC), Appearance Fidelity (AF), Imaging Quality (IQ), Temporal Consistency (TC) and Motion Smoothness (MS), all higher is better. This quantifies quality-oriented generation capabilities. We also calculate the Mean Per Joint Error (MPJPE) between the estimated 3D humans in the acting video and in the generated one, to assess the quality of the joint camera and motion control. We evaluate the Sampson Error (SE)~\cite{sampson1982fitting} for geometric consistency in presence of moving camera. \revision{We also report the 3D Consistency (3D-C) and Object Control (OC) scores from WorldScore~\cite{worldscore}: 3D-C measures multi-view coherence of the generated scene, while OC quantifies the fidelity of object appearance across frames.} In the comparison against methods with no camera control, we evaluate only motion quality with VBench metrics.

\begin{table*}[t!]
   \centering
  \caption{\textbf{Joint camera and character control.} We evaluate against Uni3C both on VBench, focusing on generation quality, and assessing control quality
   (MPJPE) and geometric consistency (SE). \revision{We also use WorldScore~\cite{worldscore} to evaluate the 3D consistency (3D-C) and object control (OC) of the generations.} We outperform in all cases Uni3C, the closest baseline in our setup.}
   \label{tab:camera_motion_results}
   \resizebox{\linewidth}{!}{
   \setlength{\tabcolsep}{8.5pt}
   \begin{tabular}{lccccccc|cccc}
   \hline
   \textbf{Model}
   & \textbf{VBench Average}$\uparrow$
   & \textbf{SC}$\uparrow$
   & \textbf{BC}$\uparrow$
   & \textbf{AF}$\uparrow$
   & \textbf{IQ}$\uparrow$
   & \textbf{TC}$\uparrow$
   & \textbf{MS}$\uparrow$
   & \textbf{MPJPE}$\downarrow$
   & \textbf{SE}$\downarrow$
   & \revision{\textbf{3D-C}$\uparrow$}
   & \revision{\textbf{OC}$\uparrow$} \\
   \hline
   Uni3C~\cite{uni3c}
   & 0.8370
   & 0.9084
   & \textbf{0.9380}
   & 0.5688
   & 0.6640
   & \textbf{0.9607}
   & 0.9821
   & 0.2121
   & 0.5665
   & \revision{0.539}
   & \revision{0.9878} \\
   \hline
   \textbf{ActCam (Ours)}
   & \textbf{0.8497}
   & \textbf{0.9212}
   & 0.9350
   & \textbf{0.5767}
   & \textbf{0.7212}
   & 0.9571
   & \textbf{0.9872}
   & \textbf{0.2087}
   & \textbf{0.4546}
   & \revision{\textbf{0.6304}}
   & \revision{\textbf{0.9953}} \\
   \hline
   \end{tabular}}
\end{table*}

\begin{table*}[t!]
 \centering
 \caption{\textbf{Static camera comparison.} We evaluate on RealisDance-Val~\cite{realisdance} with a static camera using VBench~\cite{vbench,vbenchpp}. The improved performance of ActCam compared to alternatives using 2D keypoints as conditions advocates for the superiority of our 3D-based pipeline.}
 \label{tab:main_results}
 \small
 \setlength{\tabcolsep}{10pt}
\resizebox{\linewidth}{!}{
 \begin{tabular}{lccccccc}
 \hline
 \textbf{Model} & \textbf{Average}$\uparrow$ & \textbf{SC}$\uparrow$ & \textbf{BC}$\uparrow$ & \textbf{AF}$\uparrow$ & \textbf{IQ}$\uparrow$ & \textbf{TC}$\uparrow$ & \textbf{MS}$\uparrow$ \\
 \hline
 Moore-AnimateAnyone~\cite{moore} & 83.78 & 94.65 & 94.90 & 51.56 & 66.34 & 97.16 & 98.07 \\
 HumanVid~\cite{humanvid}        & 84.68 & 93.69 & 94.94 & 55.58 & 67.45 & 97.87 & 98.52 \\
 MimicMotion~\cite{mimicmotion} & 82.27 & 92.21 & 93.60 & 52.09 & 59.67 & 97.46 & 98.61 \\
 Animate-X~\cite{animatex}      & 82.93 & 93.39 & 95.11 & 51.72 & 60.91 & 97.79 & 98.68 \\
 Hyper-Motion~\cite{hypermotion}& 84.04 & 93.58 & 94.97 & 52.97 & 65.52 & 98.19 & 99.01 \\
 UniAnimate-DiT~\cite{unianimate}& 84.29 & 94.56 & 95.44 & 52.18 & 65.52 & 98.78 & 99.24 \\
 VACE~\cite{wang2024vace}        & 85.33 & 93.56 & 95.03 & 57.81 & 70.61 & 96.74 & 98.25 \\
 Wan-Animate~\cite{wananimate}  & 84.38 & 93.06 & 94.52 & 54.47 & 66.87 & 98.42 & 98.96 \\
 SteadyDancer~\cite{steadydancer}& 85.15 & 93.48 & 95.18 & 56.80 & 68.45 & 97.99 & 99.02 \\
 \hline
 \textbf{ActCam (Ours)} & \textbf{86.47} & \textbf{95.28} & \textbf{95.83} & \textbf{58.66} & \textbf{70.83} & \textbf{98.88} & \textbf{99.34} \\
 \hline
 \end{tabular}}
\end{table*}

\subsection{Quantitative evaluation}
\label{sec:exp_bench}

\begin{figure}[b]
\centering
\includegraphics[width=\linewidth]{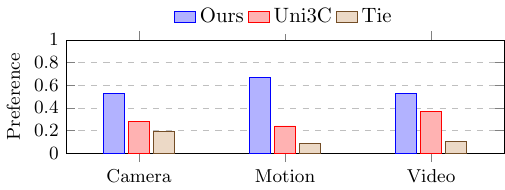}
\caption{\textbf{User study.} We compare with Uni3C on camera adherence (Camera) and motion faithfulness (Motion) with respect to the conditioning input, alongside overall visual quality (Visual). We \textit{considerably outperform} Uni3C, the closest method to ours.}
\label{fig:user_study_2afc}
\end{figure}

\subsubsection{Joint camera and motion control}
\label{sec:exp_moving}
We now present our main results. From Table~\ref{tab:camera_motion_results}, ActCam achieves higher quality/consistency scores than Uni3C and reduces motion/geometry errors under controlled camera trajectories. ActCam outperforms Uni3C in MPJPE (\textbf{0.2087} vs 0.2121) and SE (\textbf{0.4546} vs 0.5665), showcasing the superior potential of our conditioning mechanism for the moving camera and character. Interestingly, we also outperform Uni3C in the majority of VBench metrics, reporting significant improvements, especially in Subject Consistency (\textbf{0.9212}) and Imaging Quality (\textbf{0.7212}). In our zero-shot setup, we avoid loss of performance due to finetuning on restricted ad-hoc data for motion control. We propose also a qualitative comparison with Uni3C in Figure~\ref{fig:uni3c_qual_comparison}.

\paragraph{User evaluation.}
As a further comparison on joint motion and camera control, we use a two-alternative forced choice (2AFC) study on 17 users with anonymized video pairs comparing ActCam against Uni3C.
We use videos generating the same 4 camera presets as in Table~\ref{tab:camera_motion_results}, and a small subset of RealisDance-Val clips. Each trial shows two generated videos conditioned on the same inputs, alongside the reference acting video and a textual description of the camera motion. Participants then reply to questions evaluating:
(1) \textbf{Camera adherence}: ``Which video better follows the specified camera motion (viewpoint changes and stable background/parallax)?'',
(2) \textbf{Motion faithfulness}: ``Which video better matches the motion in the reference acting video (pose accuracy and smoothness)?'', and
(3) \textbf{Visual quality}: ``Which video looks more visually realistic and pleasing overall (fewer artifacts and less flicker)?''. We report the results in Figure~\ref{fig:user_study_2afc}. As visible, \textit{we considerably outperform Uni3C in all questions}, strongly suggesting that the user preference for generated videos are aligned with the performance boost reported in Table~\ref{tab:camera_motion_results}.

\subsubsection{Motion control with static camera}
\label{sec:exp_static}
We now compare against previous methods for motion control, isolating the quality of our 3D human-based motion conditioning. In Table~\ref{tab:main_results}, we show VBench metrics on RealisDance-Val. For fairness with others, we assume a static camera, and render only videos of characters in motion following the reference acting video. ActCam \textit{is consistently better} than strong baselines in this setting, improving subject/background consistency and temporal/motion metrics while maintaining high imaging quality. For example, ActCam improves TC from \emph{98.78} (UniAnimate-DiT, second best) to \emph{98.88} and AF from \emph{57.81} (VACE, second best) to \emph{58.66}, while also improving IQ from \emph{70.61} (VACE, second best) to \emph{70.83}. We attribute this result to the structural characteristics of our 3D-based conditioning: while methods based on 2D keypoints are subject to bone deformation and occlusion issues, exploiting a 3D signal regularizes subject proportions across the video.

\subsection{Qualitative evaluation}
\label{sec:exp_qual}

\begin{figure*}[p]
  \centering
  \includegraphics[width=\textwidth,height=0.45\textheight,keepaspectratio]{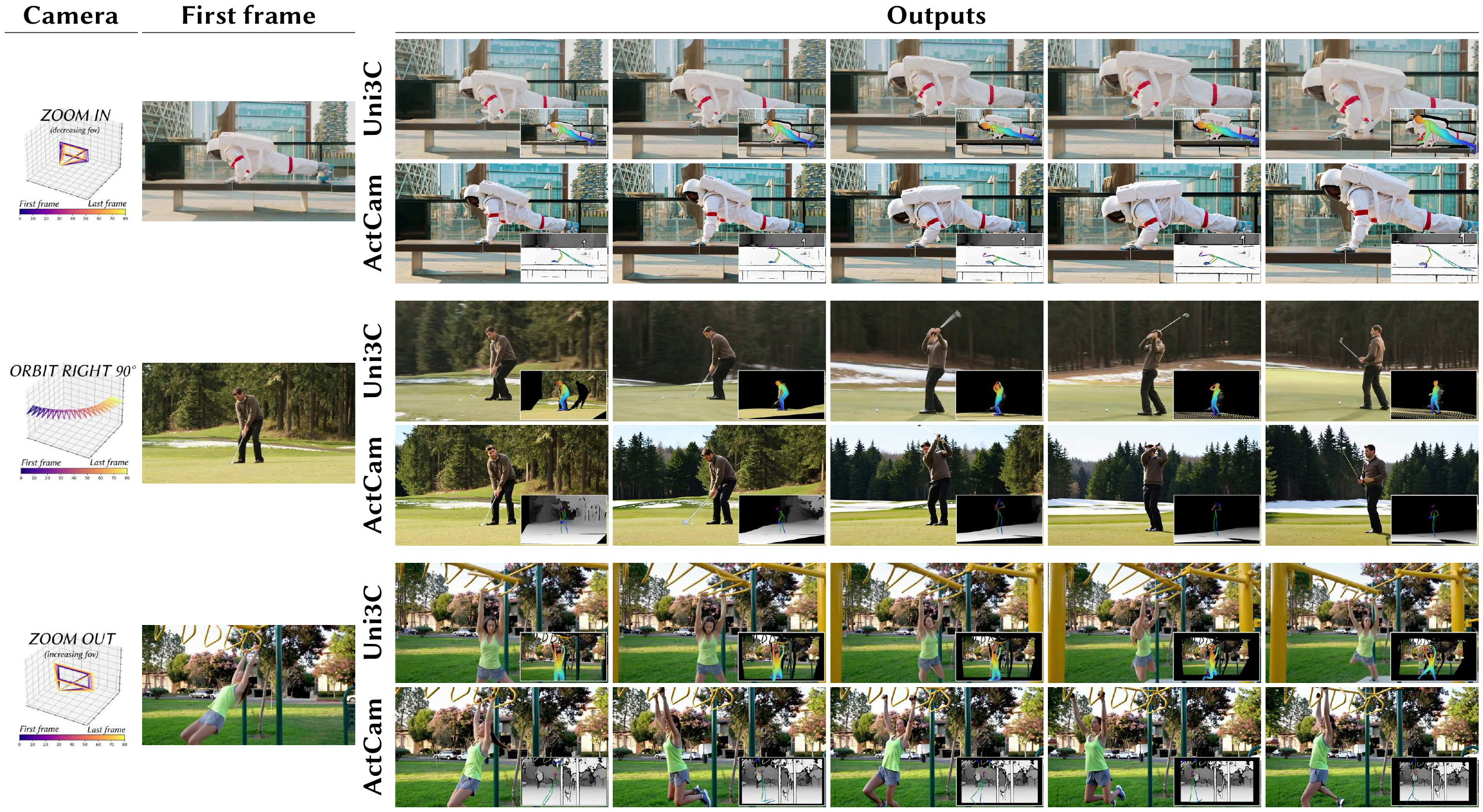}
  \caption{\textbf{Comparison with Uni3C.} Uni3C yields suboptimal camera control (top, middle) and unrealistic character motion (bottom). \revision{In the insets, a visualization of the control signal for both Uni3C and ActCam.}}
  \label{fig:uni3c_qual_comparison}

  \includegraphics[width=\textwidth,height=0.42\textheight,keepaspectratio]{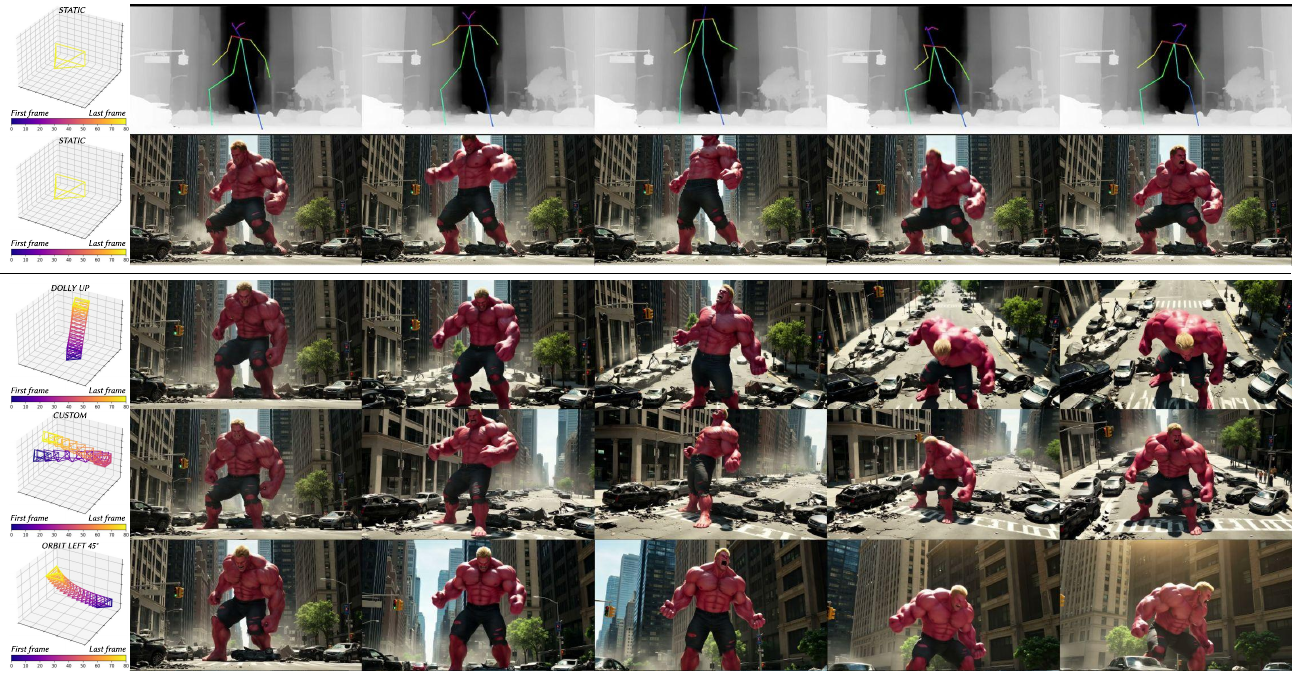}
  \caption{\textbf{Different cameras.} We first show the conditioning signal and ActCam results (top two rows). In the next three rows, we variate camera movements. As visible, the character appearance and motion remain consistent.}
  \label{fig:diff_cameras}
\end{figure*}

\begin{figure*}[p]
  \centering
  \includegraphics[width=\textwidth,height=0.2\textheight,keepaspectratio]{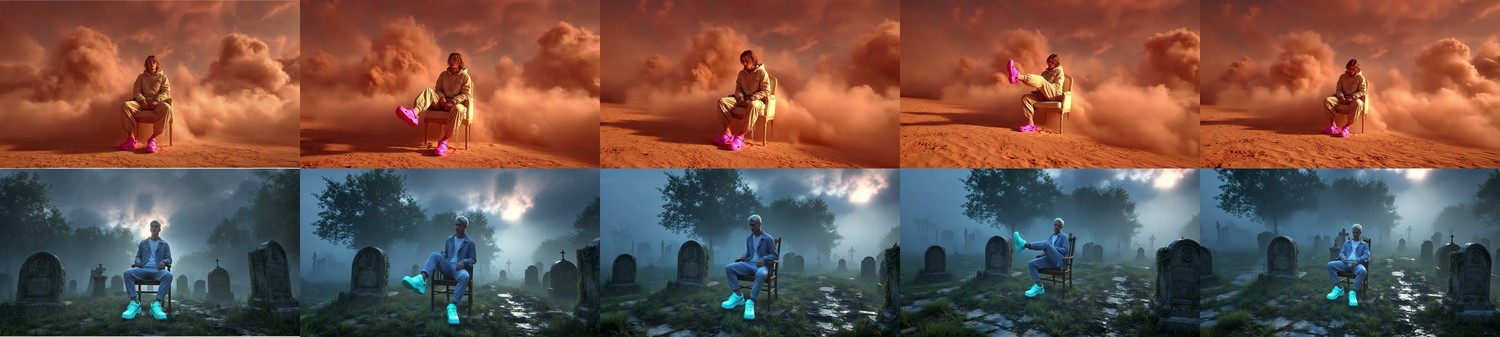}
  \caption{\textbf{Different scenes.} We display two outputs of ActCam showing the same motion rendered on two characters in different scenes, using the same camera controls.}
  \label{fig:diff_scenes_1}

  \includegraphics[width=\textwidth,height=0.2\textheight,keepaspectratio]{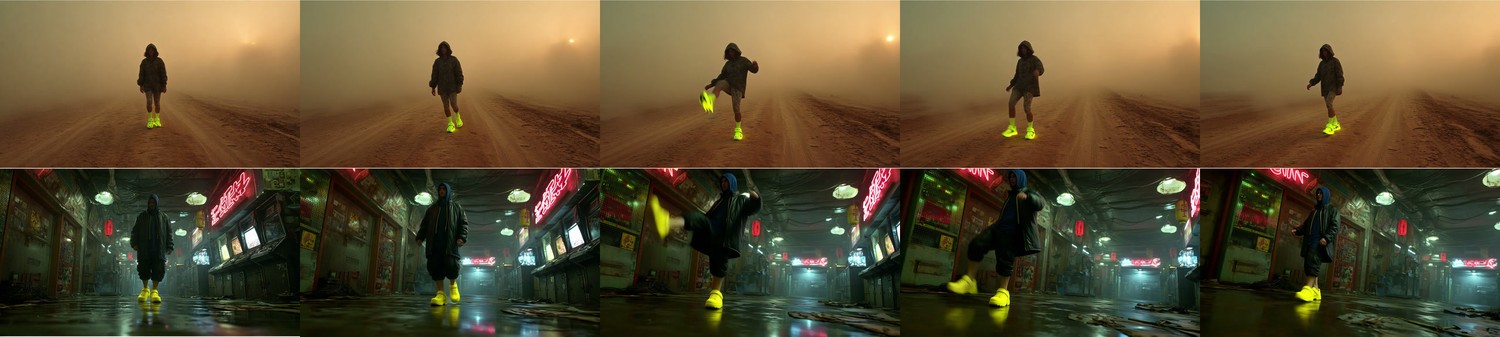}
  \caption{\textbf{Different scenes and different cameras.} To show the flexibility of our approach, we apply the same motion to two characters in different scenes, by also varying the camera control. ActCam still renders the correct motion.}
  \label{fig:diff_scenes_2}

  \includegraphics[width=\textwidth,height=0.35\textheight,keepaspectratio]{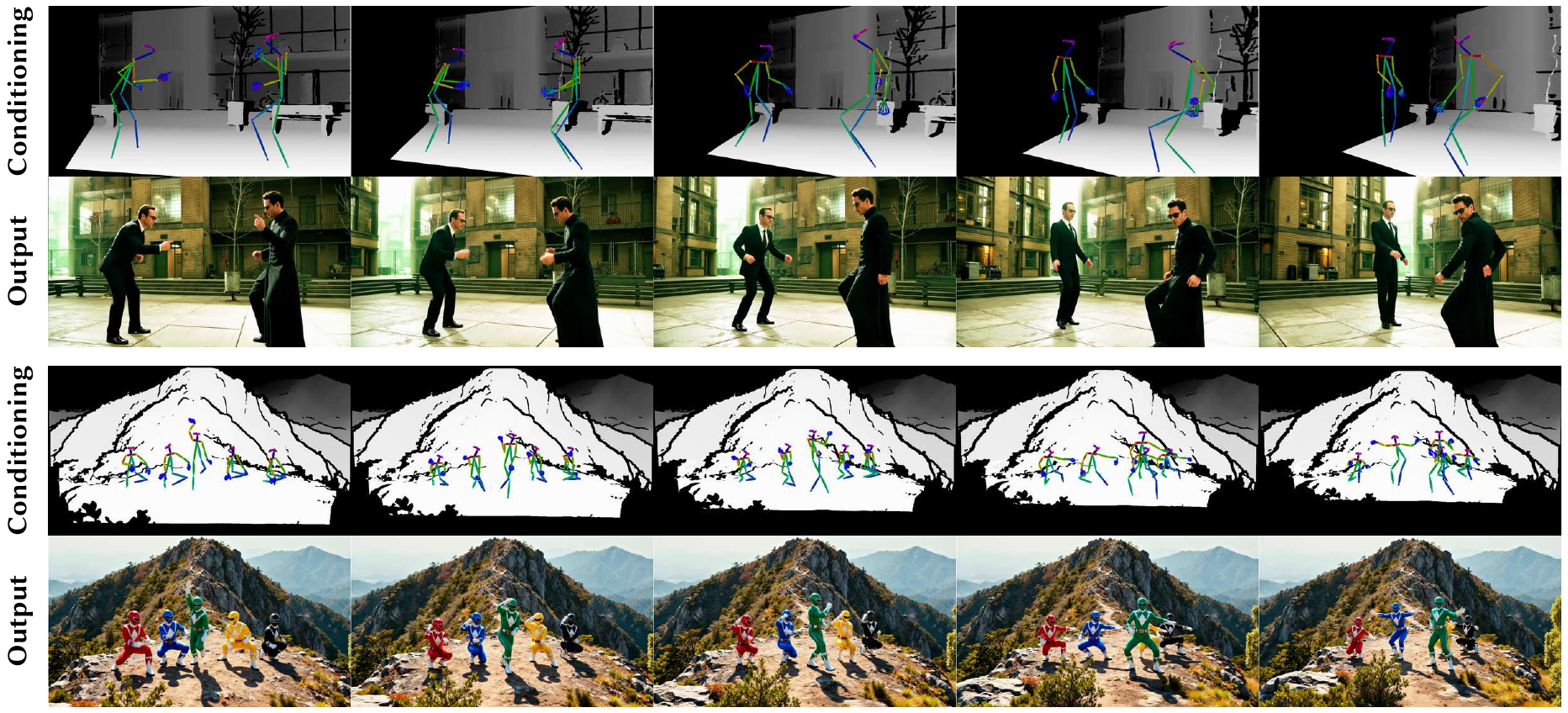}
  \caption{\revision{\textbf{Multi-character results.} ActCam handles multiple characters by applying the scene transfer and motion fitting independently per character.}}
  \label{fig:multichar}
\end{figure*}

We now report qualitative results. Besides the reported frames, we strongly suggest to visualize the supplementary video. We include multiple results showcasing the different degrees of control of ActCam. In Figure~\ref{fig:diff_cameras}, we illustrate how we can render different cameras for the same scene. Those scenarios prove the flexibility of our method across scenes, characters, and reference motion. In Figure~\ref{fig:diff_scenes_1}, we show how the same motion with the same camera can be rendered on different scenes. In Figure~\ref{fig:diff_scenes_2}, we include a camera variation, showing that even complex motion is preserved across scenarios.
\revision{Moreover, our method is adaptable to multi-character scenarios if the backbone supports it, as shown in Figure~\ref{fig:multichar}.}

\subsection{Ablation studies}
\label{sec:exp_ablate}

\begin{figure}
    \includegraphics[width=0.95\linewidth]{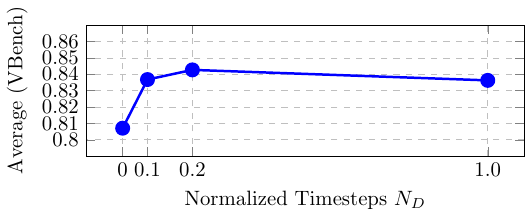}
    \caption{{\textbf{Effect of $N_D$ on VBench score.}} The figure shows the average VBench scores as a function of $N_D$, where the conditioning switches from pose+depth to pose-only. Early switching under-constrains the generation, while late switching (low $t$) can propagate depth artifacts into high-frequency details, harming results. We set an optimal $N_D=0.2$.}
    \label{fig:ablation_nd}
\end{figure}

\begin{figure}[t]
  \centering
  \includegraphics[width=\linewidth]{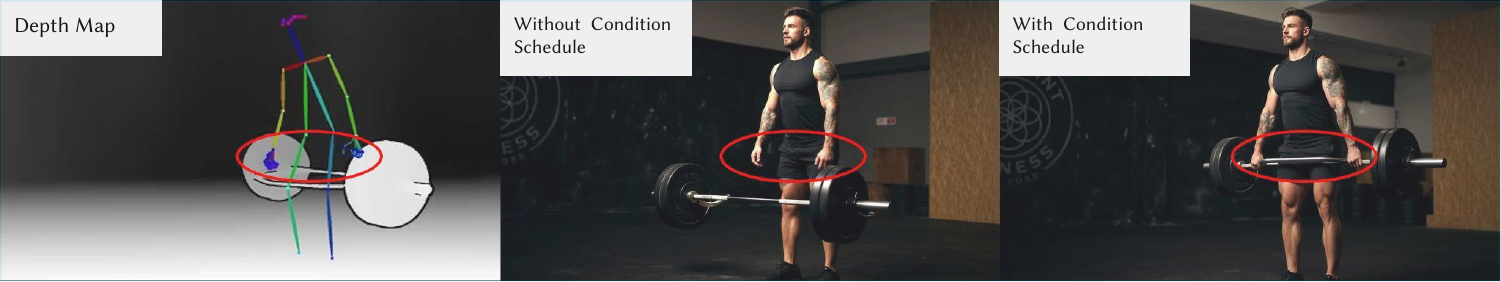}
  \caption{\textbf{Importance of conditioning schedule.} Excessive depth guidance (setting $N_D=1$) can overly constrain the scene, producing static backgrounds under camera motion (center, red circle). Instead, $N_D<1$ allows to flexibly move the barbell to follow the human motion (right).}
  \label{fig:ablation_depth_only}
\end{figure}

\begin{figure}[t]
  \centering
  \includegraphics[width=\linewidth]{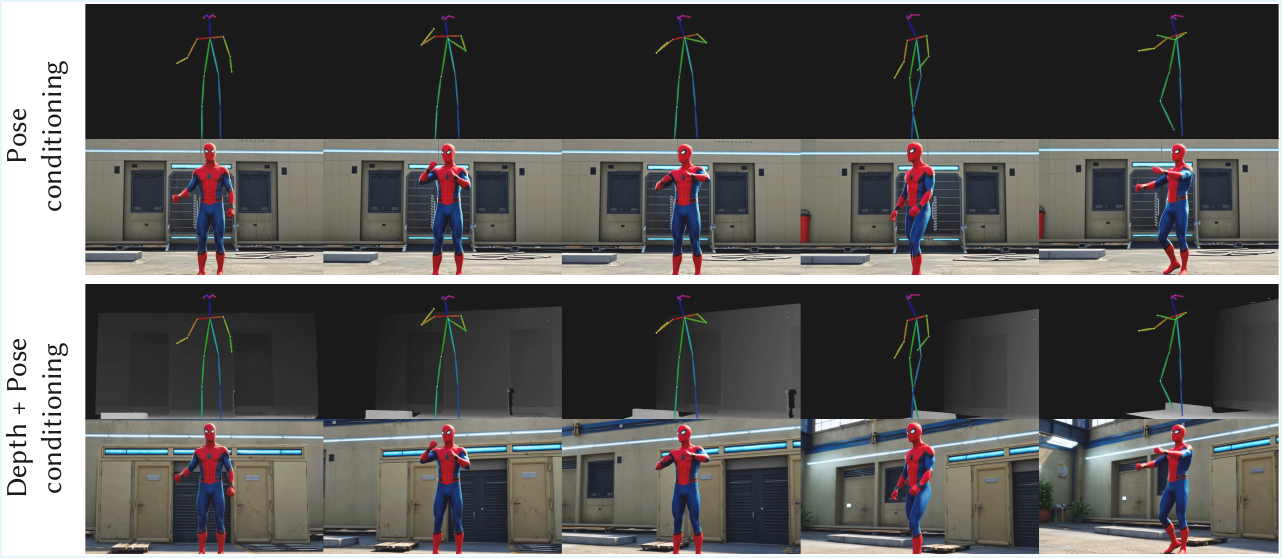}
  \caption{\textbf{Importance of depth.} Providing only pose information ($N_D=0$, top) for conditioning creates ambiguities between camera and character motion. Conversely, using depth yields the correct character and camera motions ($N_D=0.2$, bottom).}
  \label{fig:spiderman}
\end{figure}

\paragraph{Balance of depth conditioning.}
We vary the number of initial diffusion steps conditioned on both pose and depth ($N_D$) and observe a trade-off. This is reflected in VBench metrics reported in Figure~\ref{fig:ablation_nd}. We select a canonical $N_D$ that balances these effects; unless stated otherwise, we use $N_D=0.2N$ (e.g., $N_D{=}2$ when $N{=}10$). From our evaluation, introducing $N_D$ denoising iterations with depth conditioning improves environment/camera stability, improving VBench-based evaluations. However, setting $N_D$ too high can over-constrain late-stage refinement, as we show in Figure~\ref{fig:ablation_depth_only}. In there we set $N_D=1$, resulting in guidance on depth for all diffusion steps. As visible, in presence of dynamic elements or interacting objects, the rigid depth conditioning prevents motion, limiting the realism of the output scene. On the contrary, not benefiting from depth conditioning ($N_D=0$) results in ambiguities between character and camera motion, as we demonstrate in Figure~\ref{fig:spiderman}. There, providing only pose information results in a moving character on a fixed background, failing to capture adequately the camera motion.

\begin{figure}[t]
    \centering
    \newcommand{\colText}{0.03\linewidth}
    \newcommand{\colImg}{0.95\linewidth}

    \begin{minipage}[c]{\colText}
        \centering
        \rotatebox{90}{\scriptsize\textbf{No removal}}
    \end{minipage}%
    \hfill
    \begin{minipage}[c]{\colImg}
        \includegraphics[width=\linewidth]{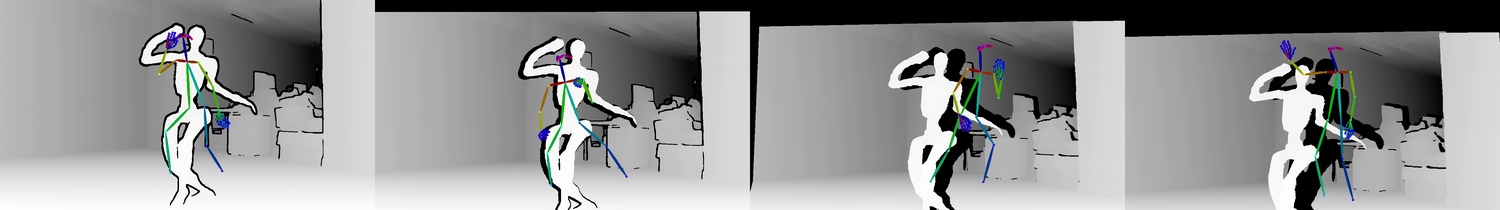}
        \par\smallskip
        \includegraphics[width=\linewidth]{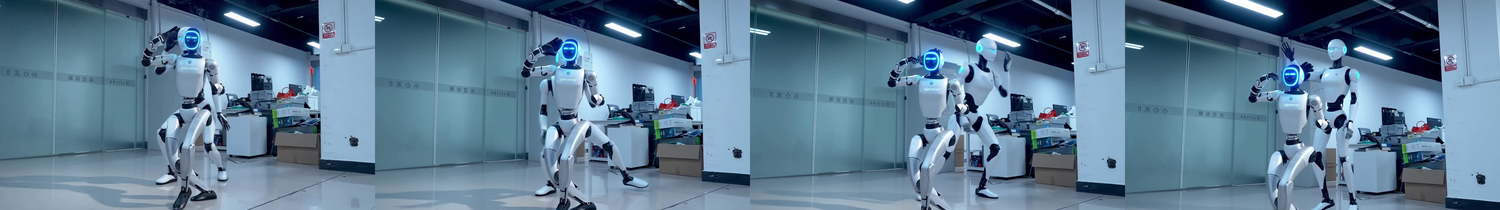}
    \end{minipage}

    \begin{minipage}[c]{\colText}
        \centering
        \rotatebox{90}{\scriptsize\textbf{With removal}}
    \end{minipage}%
    \hfill
    \begin{minipage}[c]{\colImg}
        \includegraphics[width=\linewidth]{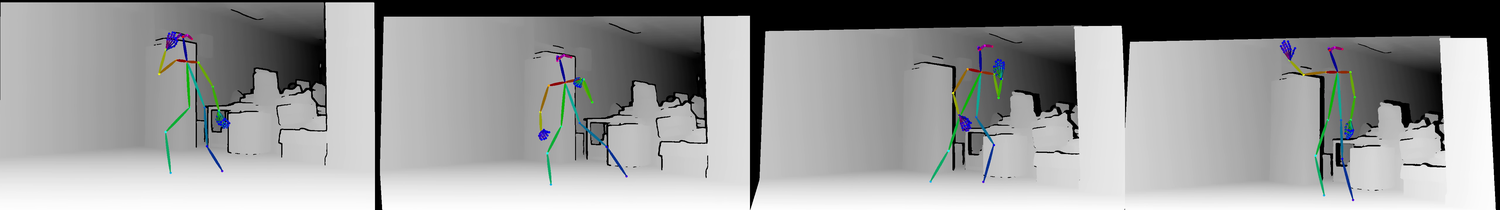}
        \par\smallskip
        \includegraphics[width=\linewidth]{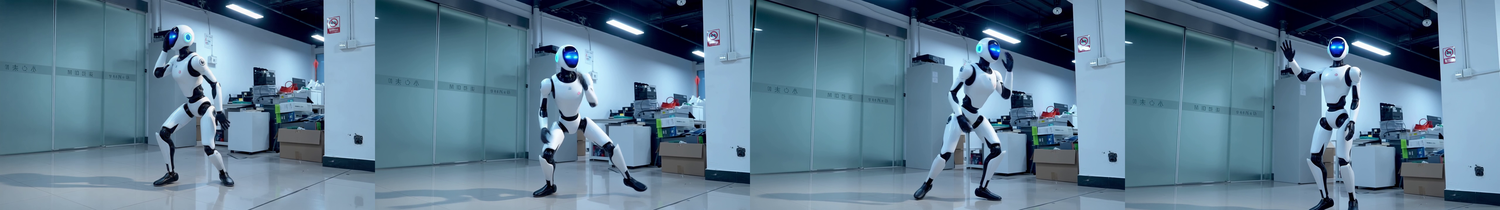}
    \end{minipage}
    \caption{\textbf{Character removal.} Without removal, the reference character is captured in the depth map, yielding duplicate subjects.}
    \label{fig:ablation_removal}
\end{figure}

\begin{figure}[t]
	\centering
	\includegraphics[width=0.85\columnwidth]{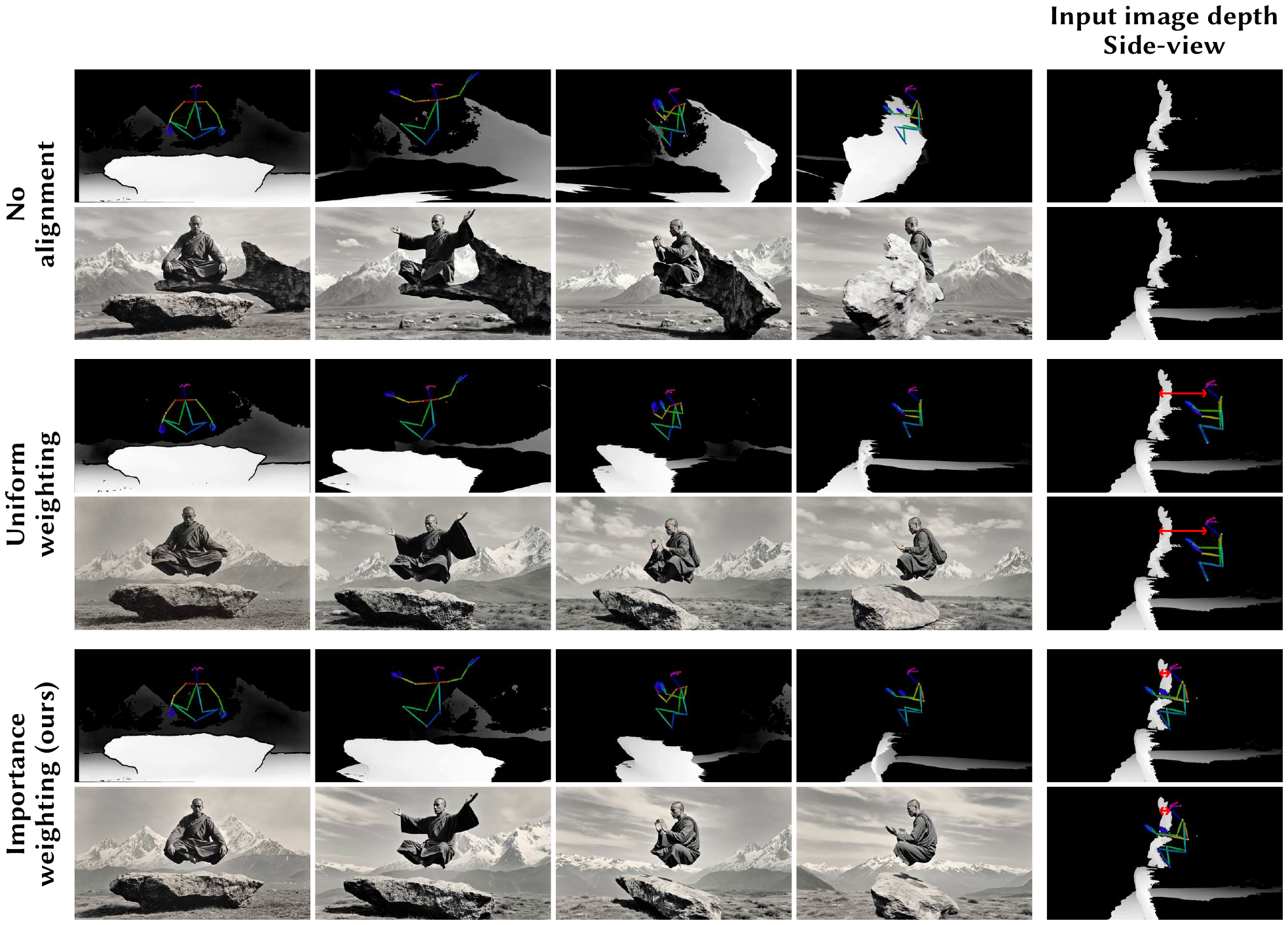}
	\caption{\textbf{Importance of scene transfer.} Without scene transfer (\emph{No alignment}), the condition does not respect 3D coherence.
		\revision{Uniform weighting improves placement but importance weighting (ours) is required to achieve best results. The red arrows (right column) show depth/positions offsets.}}
	\label{fig:ablation_depthalign}
\end{figure}

\paragraph{Reference character removal.}
As described in Sec.~\ref{sec:method:conditions}, we remove the static reference character from the reconstructed scene depth before inserting the animated character. Without this step, the static geometry of the reference character interferes with the dynamic motion conditioning in the depth map, often leading to duplicated characters. Figure~\ref{fig:ablation_removal} shows a representative example: the static character imprint in the depth signal interferes with the animated actor, yielding duplication and inconsistent compositing. This proves the importance of our design decision based on character inpainting.

\paragraph{Scene transfer.}
In Section~\ref{sec:method:conditions}, we describe how we align the composed character depth with the rendered environment depth to stabilize occlusions and prevent depth-inconsistent decomposition under difficult motions or camera changes. Without alignment, reusing the first-frame depth map can produce implausible occlusions and unstable layout when the scene contains strong depth variation or large viewpoint change. Figure~\ref{fig:ablation_depthalign} illustrates that depth alignment improves occlusion ordering and reduces layout/tearing artifacts around the actor during strong viewpoint changes.
\revision{Our importance weighting produces depth-faithful placements, as shown in Figure~\ref{fig:ablation_depthalign}.}

\section{Conclusion}
\label{sec:conclusion}

We presented ActCam, a zero-shot method for joint camera and motion control in image-conditioned video generation. ActCam constructs camera-aligned conditioning cues from a reference image, an acting video, and a target camera preset by combining target-view pose control with target-view depth-based scene guidance, while preventing static/dynamic interference through reference character removal, geometry-aware placement, and depth alignment.
To mitigate depth-induced artifacts, we introduced a two-phase conditioning schedule that uses depth only in early denoising steps to lock global structure and viewpoint changes, then relies on pose-only control for high-frequency refinement. Experiments on both static cameras and moving camera benchmarks assess the capabilities of ActCam to render high-quality videos of humans in motion. We also provided ablations and failure cases to clarify the contribution of each design choice and the remaining limitations.

\clearpage

\bibliographystyle{ACM-Reference-Format}
\bibliography{bibliography}

\end{document}